\documentclass[letterpaper, 10 pt, conference]{ieeeconf}  

\IEEEoverridecommandlockouts                              

\usepackage{graphics} 
\usepackage{epsfig} 
\usepackage{mathptmx} 
\usepackage{times} 
\usepackage{amsmath} 
\usepackage{amssymb}  
\usepackage{enumerate}
\usepackage{bm}
\usepackage{graphicx}
\usepackage{epstopdf}
\usepackage{array}
\usepackage{mathrsfs}
\usepackage{bbm}
\usepackage{multirow}

\title{\LARGE \bf
Category-level Shape Estimation for Densely Cluttered Objects
}

\author{Zhenyu Wu$^{1}$, Ziwei Wang$^{2}$, Jiwen Lu$^{2}$ and Haibin Yan$^{*1}$
	\thanks{*Corresponding author.}
	\thanks{$^{1}$Zhenyu Wu and Haibin Yan are with the School of Automation, Beijing University of Posts and Telecommunications, Beijing, 100084, China. {\tt\small \{wuzhenyu, eyanhaibin\}@bupt.edu.cn}}        
	\thanks{$^{2}$Ziwei Wang and Jiwen Lu are with the Department of Automation, Tsinghua University, and Beijing National Research Center for Information Science and Technology (BNRist), Beijing, 100084, China.  {\tt\small wang-zw18@mails.tsinghua.edu.cn, lujiwen @tsinghua.edu.cn}}
        \thanks{Code: {https://github.com/Gary3410/Shape-Estimation}}
        \thanks{This work was supported in part by the National Natural Science Foundation of China under Grant 61976623 and Grant U22B2050.}
}

\begin{document}

\maketitle
\thispagestyle{empty}
\pagestyle{empty}

\begin{abstract}

Accurately estimating the shape of objects in dense clutters makes important contribution to robotic packing, because the optimal object arrangement requires the robot planner to acquire shape information of all existed objects. However, the objects for packing are usually piled in dense clutters with severe occlusion, and the object shape varies significantly across different instances for the same category. They respectively cause large object segmentation errors and inaccurate shape recovery on unseen instances, which both degrade the performance of shape estimation during deployment. In this paper, we propose a category-level shape estimation method for densely cluttered objects. Our framework partitions each object in the clutter via the multi-view visual information fusion to achieve high segmentation accuracy, and the instance shape is recovered by deforming the category templates with diverse geometric transformations to obtain strengthened generalization ability. Specifically, we first collect the multi-view RGB-D images of the object clutters for point cloud reconstruction. Then we fuse the feature maps representing the visual information of multi-view RGB images and the pixel affinity learned from the clutter point cloud, where the acquired instance segmentation masks of multi-view RGB images are projected to partition the clutter point cloud.  Finally, the instance geometry information is obtained from the partially observed instance point cloud and the corresponding category template, and the deformation parameters regarding the template are predicted for shape estimation. Experiments in the simulated environment and real world show that our method achieves high shape estimation accuracy for densely cluttered everyday objects with various shapes. 

\end{abstract}

\section{Introduction}

Robotic packing systems \cite{liu2022ge,agarwal2020jampacker,wu2022smart,yang2021packerbot,gualtieri2021robotic,huang2022planning} play a key role in warehouse automation with the benefits of reduced uptime, high throughput, and low accident rate compared with the labor-intensive approaches. The goal of robotic packing is to stow objects into constrained space such as shipping boxes. In robotic packing systems, accurate shape estimation of objects in dense clutters is required because the planner has to obtain the shape information of all objects for packing in order to yield the optimal object arrangement in the packing boxes. For example, packing toys with different shapes leads to various placement locations and orientations, and wrong shape estimation of toys may cause packing failure due to object collision and space waste.

Encoder-decoder architectures have been widely employed in other fields \cite{wang2020learning, wang2022learning}, and have recently been utilized for object shape estimation.
RGB images \cite{stutz2018learning}, occupancy voxels \cite{wu20153d}, depth maps \cite{rock2015completing} and SDF voxels \cite{dai2017shape} of objects are embedded into the latent space with object semantics, which is followed by shape reconstruction with the decoder. Since the object size varies across instances in the same category, the object size is predicted by encoding the geometric information with the pre-defined parametric models for fine-grained shape estimation  \cite{dai2017shape,lin2021dualposenet,chen2021fs,deng2022icaps,lee2021category}. However, conventional object shape estimation methods face two challenges. First, the objects for packing are usually piled in dense clutters, and the severe occlusion among objects fails to provide informative visual clues for shape recovery. Second, the shape varies significantly for different objects in the same category, and the inaccurate shape recovery on objects with novel appearance decreases the shape estimation precision in deployment.

\begin{figure}
        \setlength{\abovecaptionskip}{0.cm}
	\includegraphics[height=3.0cm, width=8.7cm]{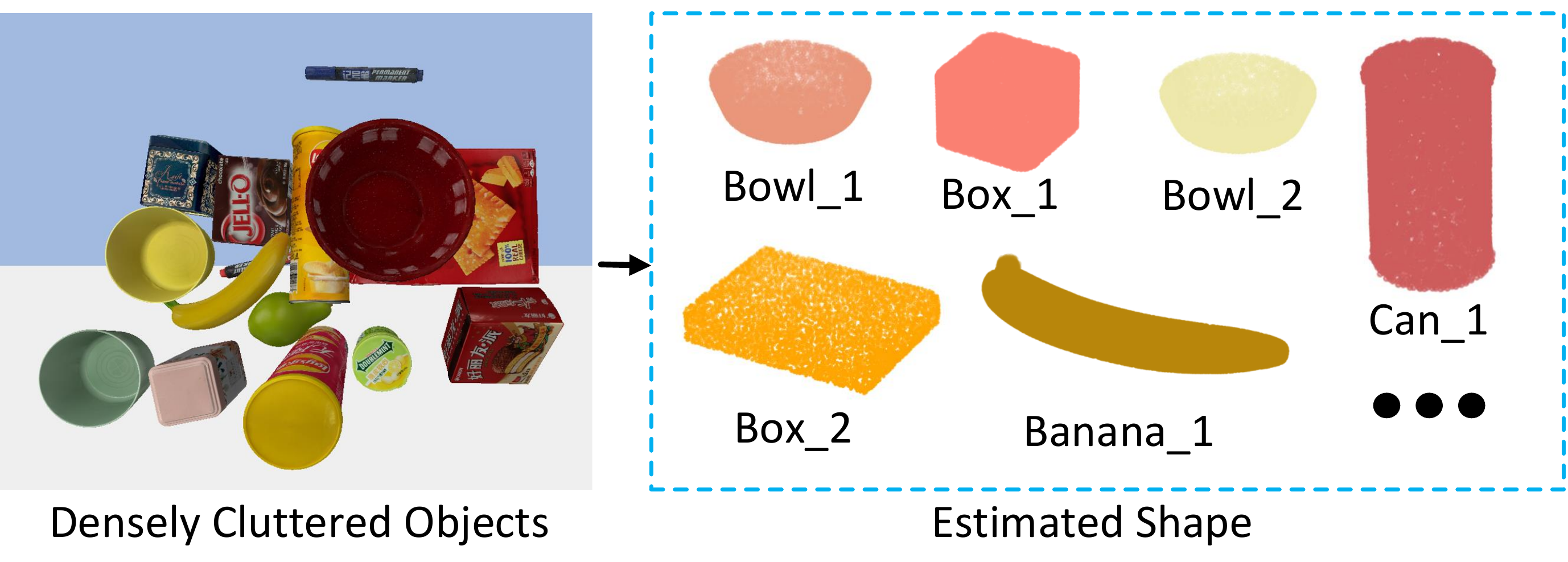}   
	\caption{An example of category-level shape estimation for densely cluttered objects.}
	\label{task}
	
	\vspace{-0.8cm}
\end{figure}

In this paper, we present a category-level shape estimation method for densely cluttered objects. Our method segments each instance in the clutter by fusing the multi-view visual information, and recovers the object shape by deforming the category templates. Hence, high segmentation precision and high generalization ability are achieved to accurately estimate the shape of all existed objects. More specifically, we collect the multi-view RGB-D images of the clutter and reconstruct the point cloud of the scene, which are utilized as the visual input of the instance segmentation module. The feature maps representing visual information of multi-view RGB images and the pixel affinity learned from the clutter point cloud are fused to generate accurate instance segmentation masks for multi-view RGB images, which are projected to the point cloud in each view for label assignment. By merging the point cloud partitions in each view with similar spatial occupancy, we obtain the observed incomplete point cloud for each object in the clutter. The observed instance point cloud and the corresponding category template are jointly utilized to regress the template deformation parameters for scale and surface transformation. Fig. \ref{task} demonstrates an example of category-level shape estimation for densely cluttered objects, where the complete point cloud of each existed instance is predicted for the object arrangement planner in robotic packing. Extensive experiments in the simulated environment and real world indicate that our framework accurately recovers the point cloud of objects in dense clutters with diverse appearances. 

\section{Related Work}
\textbf{Visual segmentation in cluttered scenes: }robotic manipulation tasks are usually challenging due to the severe occlusion in dense clutters, and object segmentation in cluttered scenes has aroused extensive interest in robotic visual perception. Existing visual segmentation for densely cluttered objects can be categorized into two types: segmentation based on RGB-D images \cite{xiang2020learning,xie2020best,xie2022rice} and point cloud \cite{dong2019ppr,xu2022fpcc}. For the first regard, robotic grasping \cite{wada2019joint,yang2020deep,schwarz2018rgb,yang2021attribute,chen2021multi} was usually guided by a visual segmentation module for the planner to generate the optimal grasp pose. In order to segment the invisible objects in the clutter for accurate visual perception, Xie \emph{et al.} \cite{xie2020best} acquired initial rough masks according to depth images and then refined the predictions with RGB images. They further mined the relationship among objects via graph neural networks to generate more accurate instance mask refinement \cite{xie2022rice}. For visual segmentation methods based on the point cloud, Dong \emph{et al.} \cite{dong2019ppr} extracted the point-wise features with the constraint that embedding of points from the same instance should be similar and vice versa, so that the clustered index in the feature space could be leveraged as the segmentation masks. Xu \emph{et al.} \cite{xu2022fpcc} inferred the geometric instance center via the learned point-wise features, and the remaining points were clustered into the closest center for segmentation. Nevertheless, the severe occlusion among objects cannot provide informative visual clues for accurate instance segmentation.

\textbf{Object shape estimation: }The goal of object shape estimation is to infer the 3D shape of objects given partial or sparse observations. Early attempts \cite{carr2001reconstruction,kazhdan2013screened} adopted surface reconstruction techniques via shape models to complete point clouds into dense surfaces. However, these methods can only model one object instance at a time with geometric priors, and the generalization ability to objects with different shapes is insufficient. Data-driven approaches \cite{rock2015completing,wu20153d} for 3D shape estimation were presented which leveraged the encoder-decoder architecture to embed the object geometry and reconstruct the full shape sequentially. Rock \emph{et al.} \cite{rock2015completing} retrieved similar objects from the database with deformation to recover the original shape. Moreover, simultaneously estimating object shape and pose \cite{avetisyan2019end,manhardt2019roi} can benefit each other due to their strong correlation.  Since object shape varies significantly across different instances in the same category, category-level shape estimation \cite{chen2021fs,lin2021dualposenet,chen2020learning,wang2019normalized,deng2022icaps,liang2021parameterized} generates the prediction with the category priors to enhance the generalization ability on unseen objects in deployment. Wang \emph{et al.} \cite{wang2019normalized} learned the canonical shape representation in the normalized object coordinate space to regress the object size regarding the category priors. However, existing methods fail to accurately recover the shape of unseen objects due to the large intra-class variation.

\begin{figure*}[t]
	\centering
	\includegraphics[height=6.5cm, width=17.5cm]{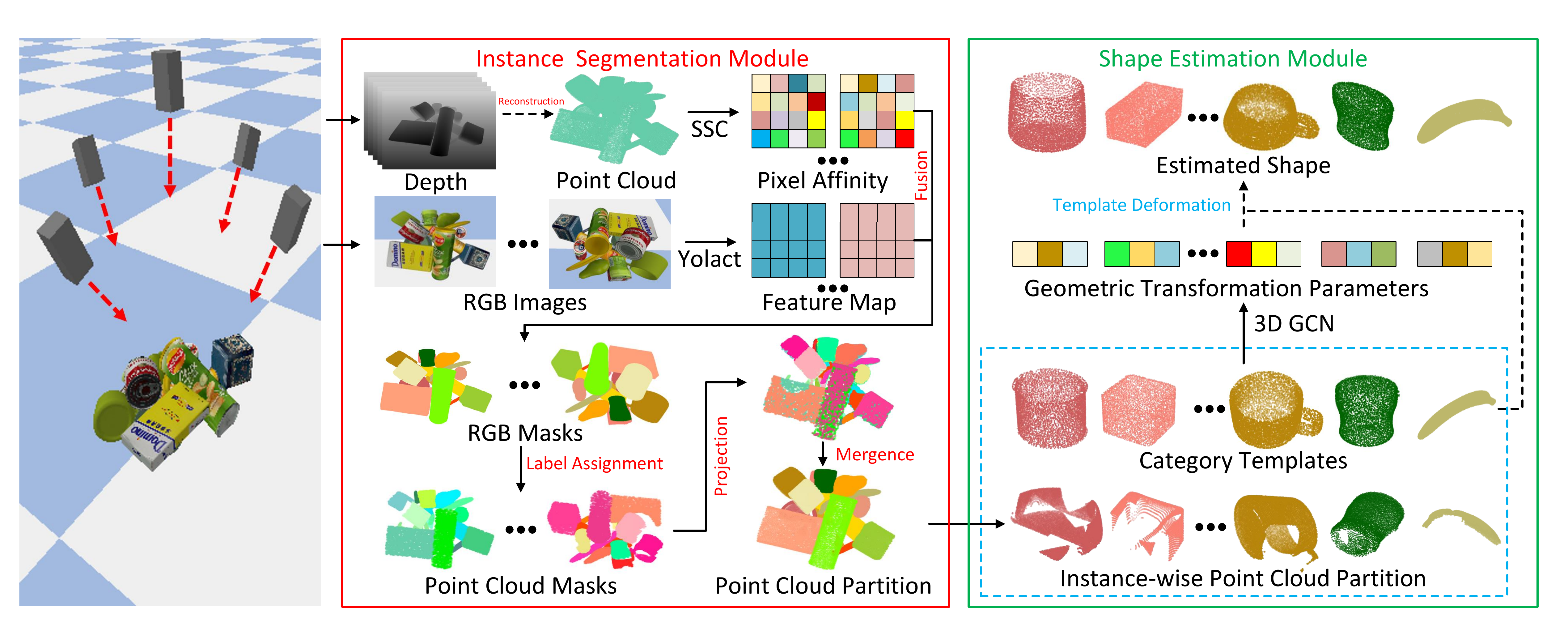}
	\vspace{-0.5cm}
	\caption{The overall pipeline of our framework, which consists of the instance segmentation module and the shape estimation module.}
	\vspace{-0.7cm}    
	\label{pipeline}
\end{figure*}

\section{Approach}
In this section, we first briefly introduce the problem of shape estimation for densely cluttered objects and the overall pipeline, and then detail the instance segmentation of object clutters by fusing the information from multi-view RGB images and point clouds. Finally, we present the category-level shape estimation for partially observed instances via template deformation.

\subsection{Problem Statement and Overall Pipeline}
The goal of shape estimation for densely cluttered objects is to predict the point cloud of every existed object in the clutter given the category template, so that the robotic packing system can yield the optimal object arrangement plan with object information. The challenges of achieving precise shape estimation are two-fold. First, the occlusion among cluttered objects disables the visual perception module to accurately recognize the object categories and segment each instance for sequential shape estimation. Second, the object shape varies significantly for instances in the same category, and objects with different shapes in deployment require a high generalization ability of the shape estimation module. To address these, we fuse the information from multi-view RGB images and clutter point cloud by passing the pixel affinity for instance segmentation, and deform the category template with diverse geometric transformation for generalizable shape estimation.

Fig. \ref{pipeline} demonstrates the overall pipeline of our framework. The object clutters are observed by one overhead and four side-view RGB-D cameras, and the side-view cameras are uniformly placed in a horizontal plane. The point cloud of the object clutters is obtained by projecting that converted from the depth image of all cameras, which is combined with the multi-view RBG images to function as the input of our framework. The pixel affinity learned from the clutter point cloud via SoftGroup \cite{vu2022softgroup} is fused into the predicted feature maps of multi-view RGB images acquired via Yolact \cite{bolya2019yolact}, which assigns the instance labels for the point cloud projected inside the mask of each view.
The pixel affinity generation process named SSC is proposed by \cite{yu2020context}.
By merging the point cloud partition across views with similar spatial occupancy, the observed point cloud for each instance is obtained. The instance-wise point cloud partition and the corresponding category template are leveraged to regress the geometric transformation parameters, where the box-cage based deformation is applied for shape estimation.

\subsection{Instance Segmentation of object clutters}
Predicting the instance-wise mask of densely cluttered objects makes significant contribution to shape estimation, because categories for different objects and instance-wise point clouds are utilized to regress the geometric transformation parameters regarding category templates. Instead of directly segmenting the clutter point cloud, we employ the instance masks of RGB images across views to assign labels to the point cloud inside the masks, as texture information significantly enhances segmentation masks for cluttered objects compared with geometry information. Since instance masks of RGB images across views may represent the same object, we should verify the object consistency across views based on predicted categories and spatial relationships to avoid false positives and negatives. The point cloud partitions from different views that share the same semantic labels and similar spatial occupancy are iteratively merged to yield the instance segmentation mask of the clutter point cloud. Let us denote the $i_{th}$ instance mask of the point cloud at the $t_{th}$ merging time as $\mathcal{P}_i^t$, the instance mask of the point cloud is updated as follows:
\begin{align}\label{mergence}
	\mathcal{P}_i^{t+1} = \mathcal{P}_i^{t}\cup\{\bm{S}_m^k|c_m^k=C_i^0, d_{ch}(\bm{S}_m^k,\mathcal{P}_i^{t})<h\},
\end{align}where $\bm{S}_m^k$ represents the $m_{th}$ point cloud partition in the $k_{th}$ view, and $c_m^k$ and $C_i^0$ respectively mean the label of $\bm{S}_m^k$ and $\mathcal{P}_i^{t}$ respectively. $d_{ch}(\bm{x}, \bm{y})$ stands for the chamfer distance between point cloud $\bm{x}$ and $\bm{y}$, and $h$ is the threshold where point clouds with chamfer distance less than $h$ are regarded to have similar spatial occupancy. Each point cloud partition is regarded as an instance at the initialization of merging, and the mergence stops to generate the instance segmentation masks for the clutter point cloud until no point cloud partition is enlarged.

Accurate instance segmentation of RGB images is crucial to precisely acquire the observed point cloud of each object for shape estimation. Severe occlusion among objects usually leads to ambiguous masks border of RGB images with incorrect predictions. Rather than directly predicting the masks of the multi-view RGB images, we fuse the pixel affinity learned via the clutter point cloud with the feature maps of RGB images to generate precise instance-wise masks for RGB images. The pixel affinity demonstrates the instance consistency among pixels, where the element in the $i_{th}$ row and $j_{th}$ column is set to one if the $i_{th}$ and $j_{th}$ pixels represent the same object and vice versa. Inspired by \cite{yu2020context}, we generate the pixel affinity $\bm{A}^k$ of the RGB image in the $k_{th}$ view based on the point cloud feature of the object clutters. For the $k_{th}$ view, the intra-class feature that fuses the information of pixels within each category and the inter-class feature which considers the visual clues from other categories are defined as follows:
\begin{align}
	&\bm{Y}_{intra}^k = \bm{A}^k\mathcal{R}(\bm{X}_{2D}^k),\quad
	\bm{Y}_{inter}^k=(\bm{1}-\bm{A}^k)\mathcal{R}(\bm{X}_{2D}^k),
\end{align}where $\mathcal{R}(\bm{X})$ means reshaping the spatial dimensions of $\bm{X}$ to match the matrix multiplication, and $\bm{1}$ is an all-one matrix with the same size as $\bm{A}^k$. Meanwhile, $\bm{X}_{2D}^k$ stands for the RGB image feature for the $k_{th}$ view. Finally, we concatenate the RGB feature, the intra-class feature, and the inter-class feature to aggregate the information for accurate instance segmentation of densely cluttered objects. Denoting the element in the $i_{th}$ row and $j_{th}$ column of $\bm{A}^k$ as $a_{ij}^k$, we aim to minimize the difference between the predicted pixel affinity matrix and groundtruth via the binary cross-entropy:
\begin{align}
	L_{ce}=-\frac{1}{KN^2}\sum_{k=1}^{K}\sum_{i,j=1}^{N}c_{ij}^k\log a_{ij}^k+(1-c_{ij}^k)\log (1-a_{ij}^k),
\end{align}where $c_{ij}^k\in\{0,1\}$ is the groundtruth label of $a_{ij}^k$, and $K$ is the number of views for visual information collection. In order to learn the correct semantic correlation for pixel affinity, we optimize the precision $L_{p}$ and recall $L_{r}$ that reveal the performance of intra-class features, and maximize the specificity $L_{s}$ that depicts the inter-class feature quality:
\begin{align}
	L_{p}=\frac{1}{K}\sum_{k=1}^{K}&\log\frac{\sum_{i,j=1}^{N}c_{ij}^ka_{ij}^k}{\sum_{i,j=1}^{N}a_{ij}^k},\quad L_{r}=\frac{1}{K}\sum_{k=1}^{K}\log\frac{\sum_{i,j=1}^{N}c_{ij}^ka_{ij}^k}{\sum_{i,j=1}^{N}c_{ij}^k},\notag\\
	&L_{s}=\frac{1}{K}\sum_{k=1}^{K}\log\frac{\sum_{i,j=1}^{N}(1-c_{ij}^k)(1-a_{ij}^k)}{\sum_{i,j=1}^{N}(1-c_{ij}^k)},
\end{align}The overall learning objective for instance segmentation considers the isolated pixel affinity correctness by binary cross-entropy and the global affinity correctness indicated by precision, recall, and specificity via the following form:
\begin{align}\label{seg_loss}
	L_{seg} = L_{ce}-\lambda(L_{p}+L_{r}+L_{s}),
\end{align}where $\lambda$ is a hyperparameter that controls the importance of global correctness in the predicted pixel affinity. By fusing the multi-view visual clues of the object clutters, we strengthen the instance segmentation accuracy of RGB images and assign precise labels for instance-wise point cloud partition for subsequent shape recovery.

\begin{figure}
	\includegraphics[height=2.6cm, width=8.5cm]{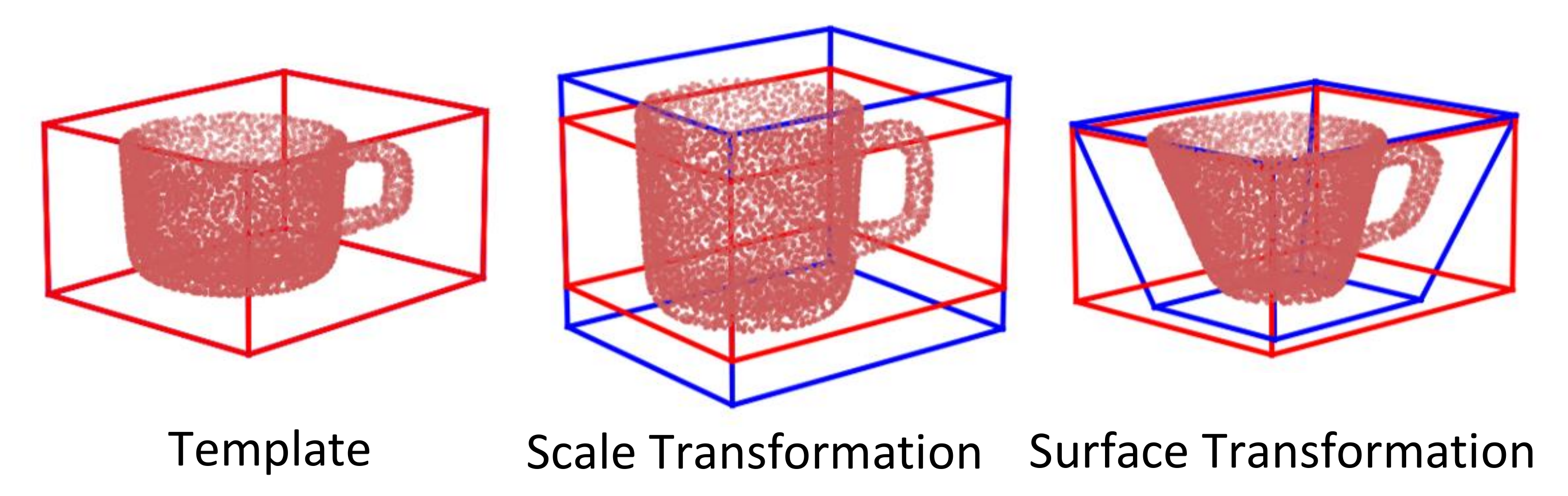}   
	\caption{Examples of scale and surface transformation for category-level templates.}
	\label{deformation}
	\vspace{-0.3cm}
\end{figure}

\begin{figure}[t]
	\centering
	\includegraphics[height=2.7cm, width=8.6cm]{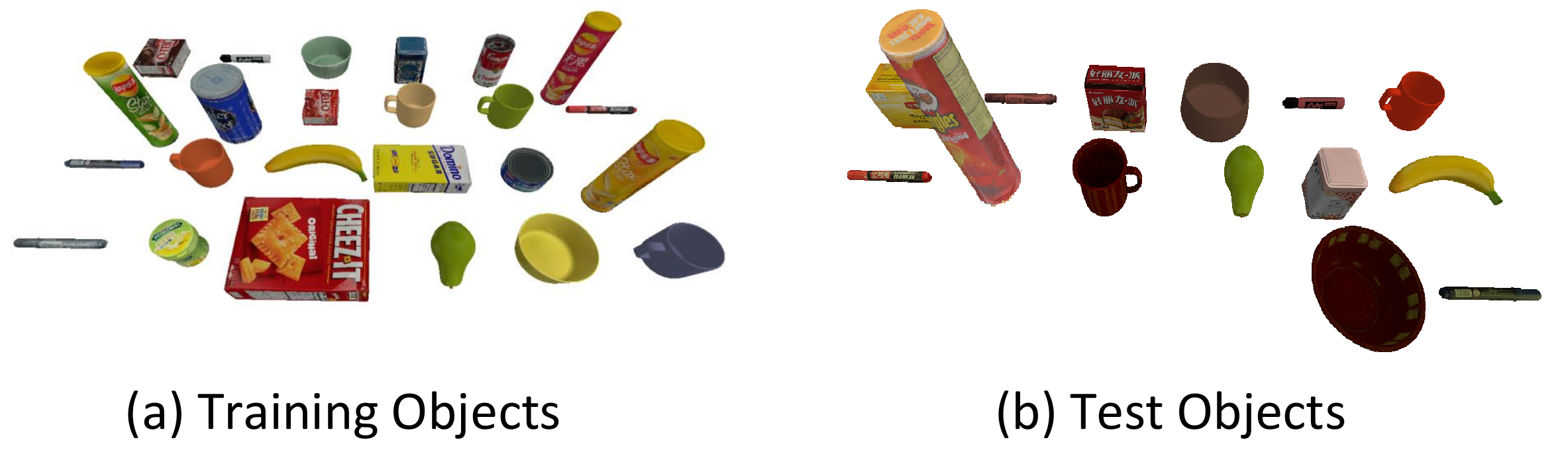}
	\vspace{-0.5cm}
	\caption{Selected objects from YCB and OCRTOC datasets for training and test in our experiments.}
	\vspace{-0.7cm}    
	\label{setting}
\end{figure}

\begin{figure}[t]
	\centering
	\includegraphics[height=2.7cm, width=8.6cm]{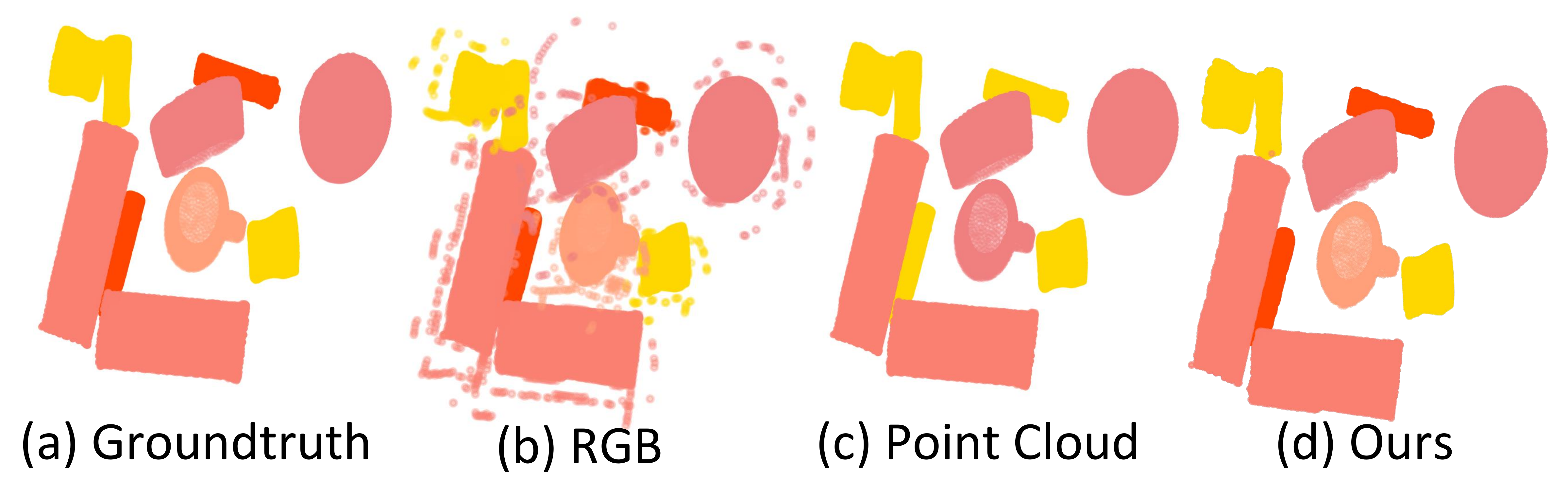}
	\caption{The visualization of instance segmentation results for (a) groundtruth, (b) RGB-only (c) point cloud-only and (d) our methods, where partitions in different colors represent various classes.}
	\vspace{-0.7cm}    
	\label{seg_vis}
\end{figure}

\begin{table*}[t]
	\centering
	\footnotesize
	\caption{The mAP and AP with different IoU thresholds of point cloud instance segmentation, where random, easy, normal and hard cases are leveraged for evaluation.}
	\vspace{-0.1cm}
	\renewcommand\arraystretch{1}
	\begin{tabular}{p{2cm}<{\centering}|p{0.8cm}<{\centering}p{0.8cm}<{\centering}p{0.8cm}<{\centering}|p{0.8cm}<{\centering}p{0.8cm}<{\centering}p{0.8cm}<{\centering}|p{0.8cm}<{\centering}p{0.8cm}<{\centering}p{0.8cm}<{\centering}|p{0.8cm}<{\centering}p{0.8cm}<{\centering}p{0.8cm}<{\centering}}
		\hline
		\multirow{2}{*}{Methods} &\multicolumn{3}{c|}{Random}&\multicolumn{3}{c|}{Easy}&\multicolumn{3}{c|}{Normal}&\multicolumn{3}{c}{Hard}\\
		\cline{2-13}
		&mAP&AP$_{25}$&AP$_{50}$&mAP&AP$_{25}$&AP$_{50}$&mAP&AP$_{25}$&AP$_{50}$&mAP&AP$_{25}$&AP$_{50}$\\
		\hline
		RGB-only \cite{bolya2019yolact}&34.16 &58.52 &51.23 &42.63 &68.70 &58.03 &33.17 &58.42 &49.21 &24.19 &52.11 &43.75 \\
		Point-only \cite{vu2022softgroup}&32.10 &64.50 &45.90 &32.50 &65.40 &47.30 &30.70 &62.40 &44.70 &27.90 &54.40 &39.00\\
		\hline
		Ours&38.75 &78.43 &58.11 &49.18 &85.43 &71.33 &38.53 &79.03 &58.22 &28.15 &74.18 &49.76\\
		\hline		
	\end{tabular}
	\vspace{-0.3cm}
	\label{tab:segmentation}
\end{table*}

\subsection{Category-level Shape Estimation of Cluttered Objects}
Acquiring the shape of each object existing in the clutter is necessary for the object arrangement planner in robotic packing systems. The point cloud partition for each instance provides visual clues with partial observation and estimating the shape of each object is equivalent to recovering the complete point cloud. Since objects from the same category share similar geometric structures \cite{vlach2016we}, we apply a box-cage based template deformation method to enhance the generalization ability of the shape estimation module on intra-class variation inspired by Fs-Net \cite{chen2021fs}. Fig. \ref{deformation} shows an example of box-cage based deformation techniques including scale and surface transformation. The predicted shape can be obtained by modifying the vertices of the template in the following:
\begin{align}
	\bm{V}_{d} = \mathbb{F}_{sur}\circ\mathbb{F}_{sca}(\bm{V}_0),
\end{align}where $\bm{V}_{d}$ and $\bm{V}_0$ respectively represent vertices of objects after and before deformation. Denoting the $i_{th}$ vertex in object vertices $\bm{V}$ as $\bm{V}^i$, the scale transformation function is defined as follows:
\begin{align}
	\mathbb{F}_{sca}(\bm{V^i})=[\alpha_xV_x^i, \alpha_yV_y^i, \alpha_zV_z^i],
\end{align}where the $V_x^i$, $V_y^i$, $V_z^i$ indicates the component in the $x$, $y$, $z$ axis for the vertex $\bm{V}^i$, and $\alpha_x$, $\alpha_y$ and $\alpha_z$ are the scaling factors for object size adjustment. The surface transformation function changes the area of the top and bottom surfaces for the box-cage in order to achieve diverse shape variations of symmetrical categories:
\begin{align}
	\mathbb{F}_{sur}(\bm{V^i})=[V_x^i+\frac{\epsilon(V_z^i-V_{z}^{\downarrow})}{V_{z}^{\uparrow}-V_{z}^{\downarrow}}V_x^i,V_y^i+\frac{\epsilon(V_z^i-V_{z}^{\downarrow})}{V_{z}^{\uparrow}-V_{z}^{\downarrow}}V_y^i, V_z^i],
\end{align}where $V_{z}^{\downarrow}$ and $V_{z}^{\uparrow}$ demonstrate the vertical coordinates of vertices with minimal and maximal $z$ value, and $\epsilon$ is the surface factor to change the ratio of top and bottom surface area for the box-cage. We construct the mesh of deformed objects with the adjusted vertices and original triangles from the template, and uniformly sample the object point cloud from the mesh for shape recovery. To regress scaling factors and surface factors in template deformation, we leverage 3D graph convolutional networks \cite{lin2020convolution} to parameterize them for each object, where instance-wise point cloud partition and vertices of corresponding category templates are employed as the input. The objective to train the shape estimation module is to minimize the Chamfer distance between the predicted point cloud and the groundtruth. By learning the correspondence pattern between the partial observation of objects and the category template, the shape estimation module reconstructs the complete point cloud for densely cluttered objects with diverse appearances.

\section{Experiments}
In this section, we conduct extensive experiments in simulated environments (Pybullet \cite{coumans2016pybullet}) and the real world to evaluate our framework. The goal of the experiment is to verify that (1) our shape estimation framework for densely cluttered objects can accurately generate complete point clouds of all existed objects, (2) the multi-view visual information fusion via pixel affinity passing significantly enhance the instance segmentation performance for object clutters, (3) deforming the category template with diverse geometric transformation according to predicted parameters strengthens the generalization ability.

\subsection{Implementation Details and Evaluation Metrics}

All objects utilized in our experiments come from the YCB dataset \cite{calli2015benchmarking} and OCRTOC dataset \cite{liu2021ocrtoc}.  We only select a subset of 24 objects for training and 14 objects for testing to construct our scenes including some generic objects such as boxes, cans, markers, sugar, bananas, pears, mugs, and bowls, where their fine-grained category names are replaced by coarse class names in category-level shape estimation. Fig. \ref{setting} visualizes the selected objects in our experiments, where most objects in the test scenarios do not appear in the training scenes. We employ the mean shape of all training objects in each category as the template. For simulated experiments, we deform the template with random parameters in scale and surface transformation for object generation to diversify the shape of instances in training and test set. We collected 1,750 and 350 RGB images from different views with pixel annotation and the corresponding clutter point cloud as the training and test set for instance segmentation module, and constructed 400 and 30 scenes which include 5-15 objects for training and testing respectively. Moreover, we prepared 24 scenes containing 5-15 objects for evaluation in real-world experiments.

Since our framework consists of the instance segmentation module and the shape estimation module, we respectively present three metrics to evaluate the above two individual modules and the overall performance on category-level shape estimation for densely cluttered objects. For instance segmentation, we leverage the mean average precision (mAP) of the point cloud masks with the IoU$\in$[0.5:0.05:0.95]. To assess the shape estimation, we utilize the Chamfer distance (CD) between the predicted and the groundtruth shape for true positive segmentation predictions. The instance segmentation module influences the precision and recall of the segmentation masks, and the shape estimation module affects the bounding box IoU between the predicted shape and the groundtruth. To measure the overall performance of our framework, we reconstruct the clutter point cloud by placing the estimated object shape with known poses and report the precision and recall of the reconstructed point cloud with various bounding box IoU thresholds. Moreover, we also provide the F1 score of the mean average precision and recall for reference.

\begin{table}[t]
	\footnotesize
	\centering
	\caption{The CD between the predicted shapes and the groundtruth for given instance-wise point cloud partitions.}
	\vspace{-0.1cm}
	\renewcommand\arraystretch{1}
	\begin{tabular}{p{2cm}<{\centering}|p{1cm}<{\centering}p{1cm}<{\centering}p{1cm}<{\centering}p{1cm}<{\centering}}
		\hline
		Deformation&Random&Easy&Normal&Hard\\
		\hline
		None &71.12 &39.71 &62.17 &147.77\\
		Scale-only&67.25 &38.00 &58.84 &139.02\\
		Surface-only&69.39 &38.82 &60.55 &143.54\\
		\hline
		Ours&63.66 &36.27 &55.67 &131.46\\
		\hline
	\end{tabular}
	\vspace{-0.6cm}
	\label{tab:shape}
\end{table}

\subsection{Simulated Experiments}
We first demonstrate the performance of the instance segmentation module with different visual information perception methods. Then we evaluate the shape estimation module with given instance-wise point clouds across various template deformations. Finally, we depict the overall performance of category-level shape estimation for densely cluttered objects.

\begin{figure}[t]
	\centering
	\includegraphics[height=3.3cm, width=8.6cm]{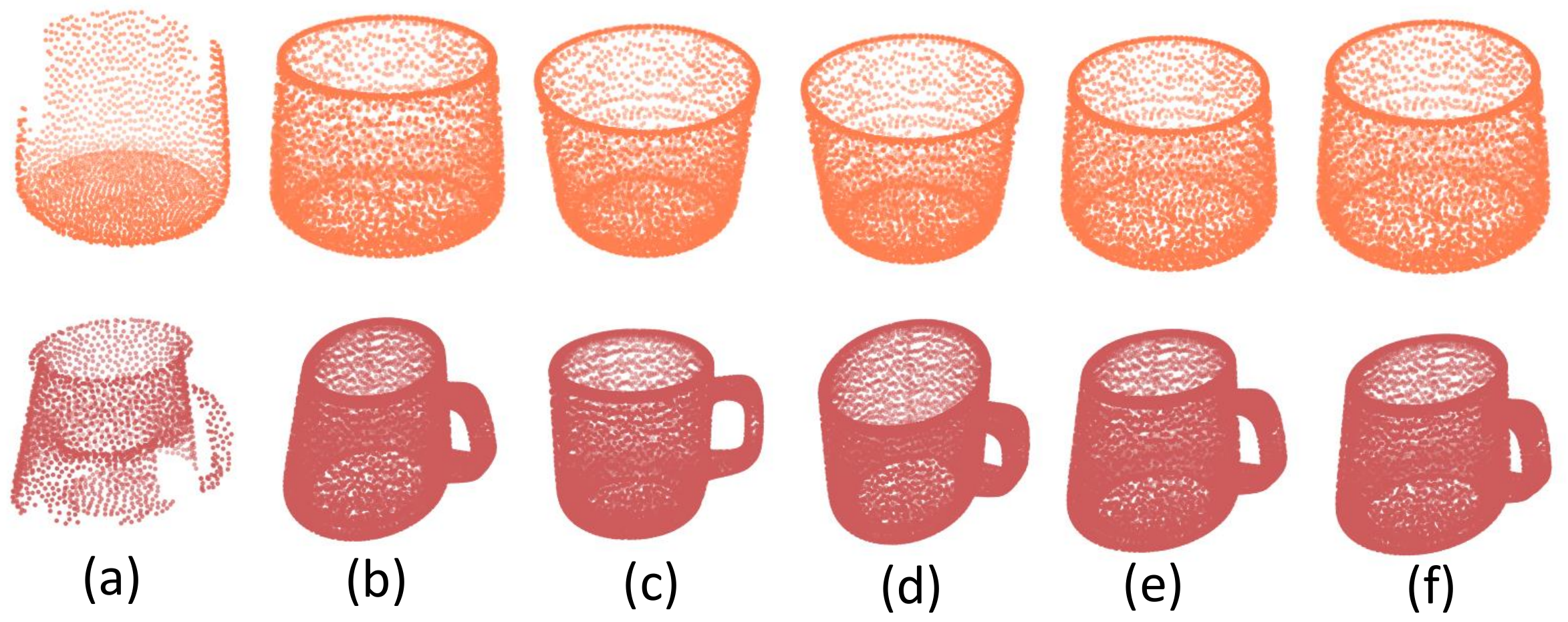}
	\caption{The visualization of (a) the observed instance point cloud for the shape estimation module, (b) the groudtruth shape, (c) the predicted shape without template deformation, (d) with only scale transformation, (e) with only surface deformation and (f) with our framework.}
	\vspace{-0.8cm}    
	\label{shape_estimation}
\end{figure}

\begin{figure*}[t]
	\centering
	\includegraphics[height=2.6cm, width=17cm]{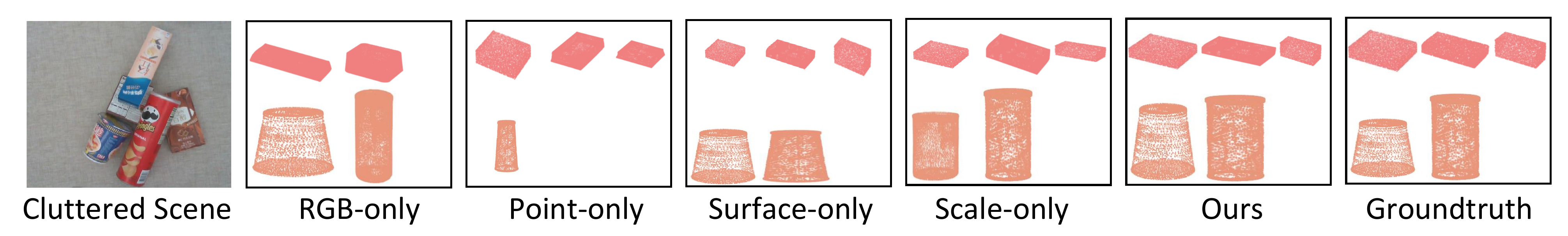}
	\vspace{-0.3cm}
	\caption{An example of the estimated shape for densely cluttered objects via different methods including RGB image based and point cloud based instance segmentation with our template deformation, surface-only and scale-only template deformation with our instance segmentation. The estimation results of our method and the groundtruth are also demonstrated.}
	\vspace{-0.2cm}    
	\label{overall_results}
\end{figure*}

\begin{table*}[t]
	\centering
	\footnotesize
	\caption{Comparison of the precision and recall with different IoU thresholds for the shape estimation of random cases. The F1 score of the mean average precision and recall is provided for reference.}
	\vspace{-0.1cm}
	\renewcommand\arraystretch{1}
	\begin{tabular}{p{1.6cm}<{\centering}|p{1.6cm}<{\centering}|p{1cm}<{\centering}p{1cm}<{\centering}p{1cm}<{\centering}p{1cm}<{\centering}|p{1cm}<{\centering}p{1cm}<{\centering}p{1cm}<{\centering}p{1cm}<{\centering}|p{1cm}<{\centering}}
		\hline
		\multicolumn{2}{c|}{Methods} &\multicolumn{4}{c|}{Precision}&\multicolumn{4}{c|}{Recall}&\multirow{2}{*}{F1 Score}\\
		\cline{1-10}
		Segmentation&Estimation&mAP&AP$_{10}$&AP$_{25}$&AP$_{50}$&mAR&AR$_{10}$&AR$_{25}$&AR$_{50}$&\\
		\hline
		RGB & None &30.25 &	50.15 &	45.84 &	10.57 &	36.38 &	46.97 &	45.52 &	13.15 &	33.03\\
		RGB &Scale&43.33 &	49.48 &	50.05 &	46.58 &	51.60 &	47.59 &	46.95 &	45.35 &	47.10\\
		RGB &Surface&40.18 &	49.53 &	47.81 &	44.57 &	48.51 &	49.40 &	47.29 &	46.37 &	43.95\\
		\hline
		Point &None&27.28 &	42.84 &	34.65 &	7.89 &	23.82 &	38.60 &	31.28 &	6.83 &	25.43 \\
		Point &Scale&41.83 &	57.10 &	45.10 &	29.58 &	40.10 &	50.93 &	40.89 &	29.24 &	40.95\\
		Point &Surface&35.79 &	49.77 &	39.39 &	21.70 &	33.02 &	44.97 &	34.93 &	20.32 &	34.35 \\
		\hline
		\multicolumn{2}{c|}{Fsnet \cite{chen2021fs}}&43.73 &	62.87 &	53.31 &	43.95 &	47.83 &	66.77 &	52.16 &	45.48 &	45.69 \\
		\multicolumn{2}{c|}{Densefusion \cite{wang2021dense}}&48.32 &	67.19 &	56.91 &	49.87 &	44.77 &	63.35 &	52.71 &	42.90 &	46.48\\
		\multicolumn{2}{c|}{Ours}&55.94 &	65.32 &	57.96 &	54.13 &	61.08 &	64.97 &	58.45 &	56.99 &	58.40\\
		\hline		
	\end{tabular}
	\vspace{-0.5cm}
	\label{tab:overall}
\end{table*}

\textbf{Results on instance segmentation: }The random clutter is constructed by dropping objects into the workspace, where the landing point for each object is selected randomly. Since the difficulty of instance segmentation is positively related to the object density, we set up the object clutters with 5, 10 and 15 objects for easy, normal and hard scenarios of instance segmentation. Table \ref{tab:segmentation} demonstrates the mAP of the instance segmentation masks of point clouds, where the baseline methods contain instance segmentation only based on multi-view RGB images \cite{bolya2019yolact} and clutter point cloud \cite{vu2022softgroup}. Compared with the method that directly segments the point cloud, we increase the mAP by 6.65\% in random cases because the texture information significantly enhances the segmentation masks for cluttered objects. Meanwhile, our framework also outperforms the baseline, which removes the pixel affinity learned from the clutter point cloud by 4.56\% in hard cases, because fusing the scene information via the clutter point cloud alleviates the segmentation errors caused by occlusion. Fig. \ref{seg_vis} visualizes the instance segmentation masks of clutter point clouds for different methods. Only leveraging the RGB images for segmentation fails to generate accurate pixel-wise masks, and methods only utilizing point cloud cannot assign the precise label to each partition.

\textbf{Results on shape estimation: }The baselines for comparison include utilizing the template as the predicted shape without deformation, with only scale transformation, and with only surface transformation. We apply the point cloud partition for each object acquired by our instance segmentation module as the input of the shape estimation module. Table \ref{tab:shape} demonstrates the Chamfer distance (CD) of different shape estimation methods, and Fig. \ref{shape_estimation} visualizes several examples of recovered shapes given the fixed partial observation of object point cloud. Our framework significantly decreases the CD compared with the baseline methods, which verifies the effectiveness of diverse geometric information in shape recovery including scale and surface transformation in shape estimation of novel objects.

\textbf{Results on shape estimation for densely cluttered objects: }By integrating the instance segmentation and the shape estimation modules, we obtain the overall performance on shape estimation for densely cluttered objects. Table \ref{tab:overall} illustrates the averaged precision and recall of the reconstructed clutter point cloud with different IoU thresholds and the mean ones with IoU from 0.1 to 0.55, where the F1 score of the mean average precision and recall is also provided for reference.
For the chosen baseline methods, we only employed the final predicted bounding boxes.
Our framework outperforms the baseline methods that combines different instance segmentation and shape estimation techniques by a sizable margin, which reveals that both instance segmentation and shape estimation are necessary to achieve practical category-level shape estimation for objects in dense clutters. Our method also achieves higher precision and recall than the state-of-the-art methods in shape estimation because of the accurate partial observation of objects and diverse template deformation.

\begin{table}[t]
	\centering
	\footnotesize
	\caption{Comparison on the precision and recall with different IoU thresholds for shape estimation of real-world experiment.}
	\vspace{-0.1cm}
	\renewcommand\arraystretch{1}
	\begin{tabular}{p{1.5cm}<{\centering}|p{0.8cm}<{\centering}p{0.8cm}<{\centering}|p{0.8cm}<{\centering}p{0.8cm}<{\centering}|p{1cm}<{\centering}}
		\hline
		\multirow{2}{*}{Methods} &\multicolumn{2}{c|}{Precision}&\multicolumn{2}{c|}{Recall}&\multirow{2}{*}{F1 Score}\\
		\cline{2-5}
		&mAP &AP$_{25}$ &mAR &AR$_{25}$&\\
		\hline
		RGB-only &46.26 &55.93 &45.63 &51.26 &45.94 \\
		Point-only &38.43 &40.61 &42.48 &49.97 &40.35 \\
		Scale-only &50.43 &	56.15 &	41.43 &	46.50 &	45.49 \\
		Surface-only &40.14 &	49.35 &	36.11 &	40.58 &	38.02 \\
		\hline
		Ours &54.26 &59.82 &50.15 &53.22 &52.12 \\
		\hline		
	\end{tabular}
	\vspace{-0.5cm}
	\label{tab:Real_sense}
\end{table}

\subsection{Real-world Experiments}

Fig. \ref{overall_results} shows several quantitative examples for estimating the shape of all objects in the dense clutters. RGB-only and point-only represent the methods only leveraging the RGB images and point cloud for instance segmentation following our template deformation techniques. Surface-only and scale-only depict the approaches that utilize our instance segmentation module following the surface and scale transformation for template deformation respectively. Compared with the RGB-only and point-only methods, our framework accurately segments each instance without missing objects because of global information fusion. The surface-only and scale-only methods cannot precisely estimate the shape of each instance due to the limited geometric transformation of templates. For example, the instant noodle bucket in Fig. \ref{overall_results} uses the same bucket template as the potato chip bucket applies in template deformation, and our framework can effectively estimate the shape of the instant noodle bucket with the help of surface transformation even their shapes differ obviously. Table \ref{tab:Real_sense} illustrates the average precision and recall with different IoU thresholds of our framework in real-world experiments. We also provide the F1 score of the mean average precision and recall for reference. Our framework outperforms baseline methods, which further verifies the effectiveness of both our instance segmentation and shape estimation modules in practical scenarios. The difficulties in the simulated and real-world scenarios are similar due to the same object number setting. The mAP acquired in the real-world experiment is only 1.68\% lower than that in the simulated environment, which reveals the high generalization ability of our method to real-world cluttered object shape estimation.

\section{Conclusion and Future Work}
In this paper, we have presented a category-level shape estimation framework for densely cluttered objects. We collect the multi-view RGB-D images of the object clutters and reconstruct the point cloud of the whole scene for visual clue representation. The feature maps of multi-view RGB images and the pixel-wise similarity learned from the clutter point cloud are fused via affinity passing for accurate instance segmentation of RGB images, which assigns correct labels for point clouds of each view to acquire the instance point cloud with mergence. The correspondence pattern between the instance-wise point cloud partition and the category template is extracted to predict the parameters of geometric transformation regarding templates for shape estimation. Extensive experiments in the simulated environment and real world demonstrate the effectiveness of the proposed method. In future work, we plan to reduce the computational and storage complexity of pixel affinity prediction and diversify the geometric transformation for template deformation.


{
	\bibliographystyle{ieee}
	\bibliography{egbib}
}

\end{document}